\documentclass[conference]{IEEEtran}
\IEEEoverridecommandlockouts
\usepackage{cite}
\usepackage{amsmath,amssymb,amsfonts}
\usepackage{algorithm,algorithmic}
\usepackage[dvips]{graphicx}
\usepackage{textcomp}
\usepackage{xcolor}

\usepackage{subfigure}
\usepackage{url}
\usepackage{listings}

\newcommand{\bvec}[1]{\mbox{\boldmath $#1$}}

\def\BibTeX{{\rm B\kern-.05em{\sc i\kern-.025em b}\kern-.08em
    T\kern-.1667em\lower.7ex\hbox{E}\kern-.125emX}}
\begin{document}

\title{Adaptive Structural Learning of Deep Belief Network for Medical Examination Data and \\Its Knowledge Extraction by using C4.5 \\
\thanks{\copyright 2018 IEEE. Personal use of this material is permitted. Permission from IEEE must be obtained for all other uses, in any current or future media, including reprinting/republishing this material for advertising or promotional purposes, creating new collective works, for resale or redistribution to servers or lists, or reuse of any copyrighted component of this work in other works.}
}

\author{\IEEEauthorblockN{1\textsuperscript{st} Shin Kamada}
\IEEEauthorblockA{\textit{Graduate School of Information Sciences} \\
\textit{Hiroshima City University}\\
3-4-1, Ozuka-Higashi, Asa-Minami-ku,\\
Hiroshima, 731-3194, Japan\\
Email: da65002@e.hiroshima-cu.ac.jp}
\and
\IEEEauthorblockN{2\textsuperscript{nd} Takumi Ichimura}
\IEEEauthorblockA{\textit{Faculty of Management} \\ \textit{and Information Systems} \\
\textit{Prefectural University of Hiroshima}\\
1-1-71, Ujina-Higashi, Minami-ku,\\
Hiroshima, 734-8559, Japan\\
Email: ichimura@pu-hiroshima.ac.jp}
\and
\IEEEauthorblockN{3\textsuperscript{rd} Toshihide Harada}
\IEEEauthorblockA{\textit{Faculty of Health and Welfare} \\
\textit{Prefectural University of Hiroshima}\\
1-1-71, Ujina-Higashi, Minami-ku,\\
Hiroshima, 734-8559, Japan\\
Email: hartoshi@pu-hiroshima.ac.jp}
%\and
%\and
%\and
}

\maketitle

\begin{abstract}
Deep Learning has a hierarchical network architecture to represent the complicated feature of input patterns. The adaptive structural learning method of Deep Belief Network (DBN) has been developed. The method can discover an optimal number of hidden neurons for given input data in a Restricted Boltzmann Machine (RBM) by neuron generation-annihilation algorithm, and generate a new hidden layer in DBN by the extension of the algorithm. In this paper, the proposed adaptive structural learning of DBN was applied to the comprehensive medical examination data for the cancer prediction. The prediction system shows higher classification accuracy (99.8\% for training and 95.5\% for test) than the traditional DBN. Moreover, the explicit knowledge with respect to the relation between input and output patterns was extracted from the trained DBN network by C4.5. Some characteristics extracted in the form of IF-THEN rules to find an initial cancer at the early stage were reported in this paper.
\end{abstract}


\section{Introduction}
Recently, various deep learning methods to learn large volume data have been developed in many fields and have been utilized to make our life more abundant \cite{Bengio09, Quoc12, webmarket2016}. Especially deep learning techniques of image diagnosis for medical data have been extensively and rapidly studied \cite{Russakovsky15}. Their new architectures \cite{AlexNet, GoogLeNet, VGG16, ResNet} have been developed for image recognition.

In our study, we have proposed the adaptive structural learning of Deep Belief Network (DBN) \cite{Kamada18_Springer}. The method is the hierarchical structure of adaptive structural learning of Restricted Boltzmann Machine (RBM) \cite{Hinton06,Hinton12} which has the self-organized function by generating / deleting hidden neurons during learning. The number of RBMs is also automatically defined by the layer generation method which monitors the error and variances of some parameters \cite{Kamada16_SMC, Kamada16_ICONIP, Kamada16_TENCON}. The developed method shows the highest classification capability of image recognition of the benchmark data set such as MNIST \cite{LeCun98a}, CIFAR-10, and CIFAR-100 \cite{CIFAR10}. The classification ratios for data sets were almost 100\% for training cases and 99.5\%, 97.4\%, and 81.2\% for test cases respectively at the time of writing this paper \cite{Kamada18_Springer}. 

In this paper, the prediction system for cancer is developed on the tremendous comprehensive medical examination data \cite{Kanhokyo} by the adaptive structural learning of DBN. The medical examination data consists of about 100,000 records which have not only blood tests and inquiries but also medical images such as chest X-ray, CT, and MRI images. The training results of our method showed higher classification performance than the traditional RBM and DBN.

Moreover, the explicit knowledge represented as IF-THEN rules is extracted from the trained DBN network by using C4.5 \cite{Quinlan96}. The knowledge extraction which transforms implicit knowledge in deep learning to explicit knowledge is one of the important subjects \cite{Kamada16_IWCIA, Kamada17_IJCIS, Kamada17_SMC}. The extracted IF-THEN rules show some interesting features for the relation between blood tests and the probability of cancer without medical images such as CT or MRI. There are several rules that fail to notice due to the symptoms at the early stage, but can lead to early detection. These abnormalities in only blood test items can not be said to be cancer, but they will be medically considered cases at the early stage \cite{alb, a6}. We investigated the trained adaptive structural DBN for medical comprehensive examination data and then we report some interesting rules.

The remainder of this paper is organized as follows. In section \ref{sec:adaptive_dbn}, the adaptive structural learning of DBN is explained. Section \ref{sec:extract_knowledge} gives the proposed knowledge extraction method of the DBN. Section \ref{subsec:exe_cancer_prediction} and \ref{subsec:exe_rule} show the simulation result and the extracted knowledge, respectively. In Section \ref{sec:conclusion}, we give some discussions to conclude this paper.

\section{Adaptive Learning Method of Deep Belief Network}
\label{sec:adaptive_dbn}
\subsection{Restricted Boltzmann Machine}
A RBM \cite{Hinton12} is a stochastic unsupervised learning model. The network structure of RBM consists of two kinds of binary layers: one is a visible layer $\bvec{v} \in \{0, 1 \}^{I}$ with the parameter $\bvec{b} \in \mathbb{R}^{I}$ for input patterns, and the other is a hidden layer $\bvec{h} \in \{0, 1 \}^{J}$ with the parameter $\bvec{c} \in \mathbb{R}^{J}$ to represent the feature of input space. $I$ and $J$ are the number of visible and hidden neurons, respectively. The connection between a visible neuron $v_{i}$ and a hidden neuron $h_j$ is represented as the weight $W_{ij}$. There are not any connections among the neurons in the same layer. A RBM tries to minimize the energy function $E(\bvec{v}, \bvec{h})$ in training, which is defined as follows, with the learning parameters $\bvec{\theta}=\{\bvec{b}, \bvec{c}, \bvec{W} \}$. 

\begin{equation}
E(\bvec{v}, \bvec{h}) = - \sum_{i} b_i v_i - \sum_j c_j h_j - \sum_{i} \sum_{j} v_i W_{ij} h_j ,
\label{eq:energy}
\end{equation}
\vspace{-2mm}
\begin{equation}
p(\bvec{v}, \bvec{h})=\frac{1}{Z} \exp(-E(\bvec{v}, \bvec{h})),
\label{eq:prob}
\end{equation}
\vspace{-2mm}
\begin{equation}
Z = \sum_{\bvec{v}} \sum_{\bvec{h}} \exp(-E(\bvec{v}, \bvec{h})),
\label{eq:z}
\end{equation}
where $b_i$ and $c_j$ are the parameters for visible neuron $v_i$ and hidden neuron $h_j$, respectively. $W_{ij}$ is a parameter (weight) between $v_i$ and $h_j$. Eq.~(\ref{eq:prob}) shows the probability of $\exp(-E(\bvec{v}, \bvec{h}))$. $Z$ is calculated by summing energy for all possible pairs of visible and hidden vectors. The parameters $\bvec{\theta}=\{\bvec{b}, \bvec{c}, \bvec{W} \}$ are optimized for given input data by partial derivative of $p(\bvec{v})$. A standard way of estimating the parameters of a statistical model $p(\bvec{v})$ is maximum likelihood estimation.

Contrastive Divergence (CD) \cite{Hinton02} is the most popular RBM learning method as a fast algorithm of Gibbs sampling method. The sampling is implemented for the states on Markov chain by Monte Carlo. Since the $\exp(-E(\bvec{v}, \bvec{h}))$ is discrete values on $\bvec{v}$ and $\bvec{h}$, we give some discussions of RBM learning with CD binary sampling under the Lipschitz continuous \cite{Carlson15, Kamada16_SMC}. In previous research, we investigated the change of gradients for the parameters $\bvec{\theta}=\{\bvec{b}, \bvec{c}, \bvec{W} \}$ during the learning phase, then the two kinds of parameters $\bvec{c}$ and $\bvec{W}$ were influenced on the convergence situation of RBM except the parameter $\bvec{b}$ because the gradient for parameter $\bvec{b}$ may be affected by various features of input patterns.

\begin{figure}[bt]
\centering
\includegraphics[scale=0.8]{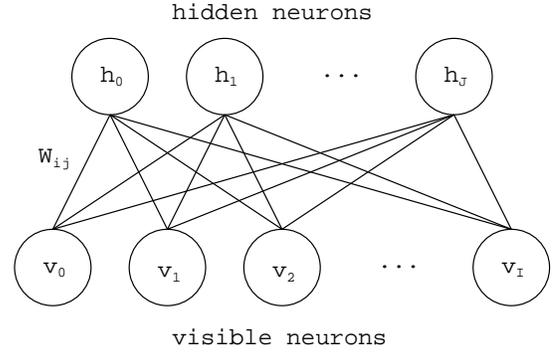}
%\vspace{-3mm}
\caption{Network Structure of RBM}
\label{fig:rbm}
\end{figure}

\begin{figure*}[tb]
\centering
\includegraphics[scale=0.8]{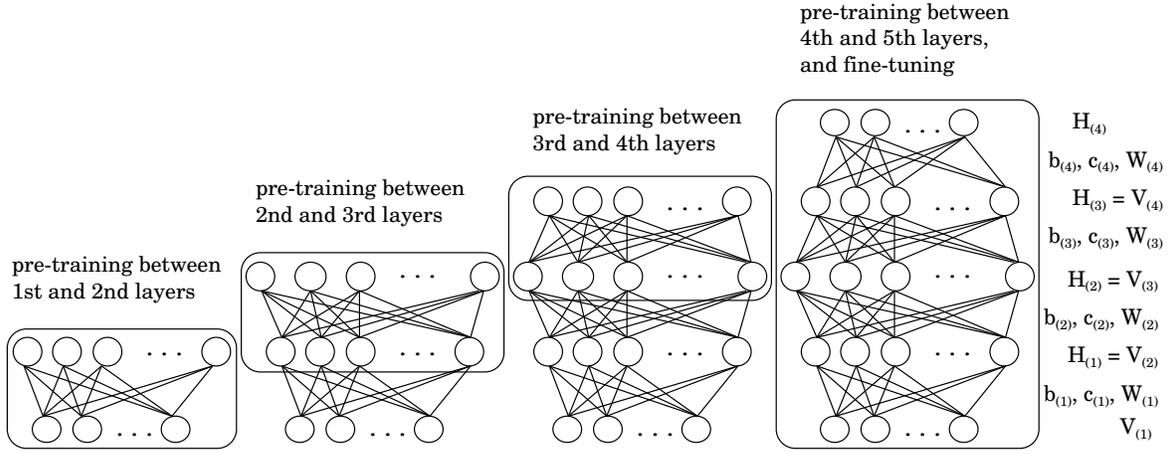}
%\vspace{-3mm}
\caption{Network structure of DBN}
\label{fig:dbn}
\end{figure*}

\subsection{Deep Belief Network}
\label{subsec:dbn}
Deep Belief Network (DBN) \cite{Hinton06} is a stacking box for stochastic unsupervised learning by hierarchically building several pre-trained RBMs. The hierarchical network structure of DBN can be constructed by building two or more pre-trained RBMs. The output patterns of hidden neurons at the $l$-th RBM can be seen as the next input at the $(l+1)$-th RBM. The conditional probability of a hidden neuron $j$ at the $l$-th RBM is defined by Eq.(\ref{eq:prob_dbn}).
\begin{equation}
\label{eq:prob_dbn}
p(h_j^{l} = 1 | \bvec{h}^{l-1})= sigm(c^{l}_j + \sum_{i}W^{l}_{ij} h^{l-1}_{i}),
\end{equation}
where $c^{l}_j$ and $W^{l}_{ij}$ are the parameters for a hidden neuron $j$ and the weight at the $l$-th RBM, respectively. $\bvec{h}^{0} = \bvec{v}$ is the given input data. When the trained DBN takes a supervised learning for a classification task, an output layer is added to the last layer, then the output probability $y_k$ for a category $k$ is calculated by Softmax at the output layer as shown in Eq.(\ref{eq:softmax}). 
\begin{equation}
\label{eq:softmax}
y_k = \frac{\exp(z_{k})}{\sum^{M}_{j} \exp(z_j)},
\end{equation}
where $z_{j}$ is an output pattern of a hidden neuron $j$ at the output layer. $M$ is the number of output neurons. The error between the output $y_k$ and the teacher signal is minimized.

\subsection{RBM's Neuron Generation / Annihilation Algorithm}
\label{subsec:adaptive_rbm}
Generally, the decision of an optimal architecture of deep learning network to given training data is one of eternal problems for the network designer, since trial and error for finding its parameter set is required to get higher classification capability.

\begin{figure}[tbp]
\begin{center}
\subfigure[Neuron Generation]{\includegraphics[scale=0.5]{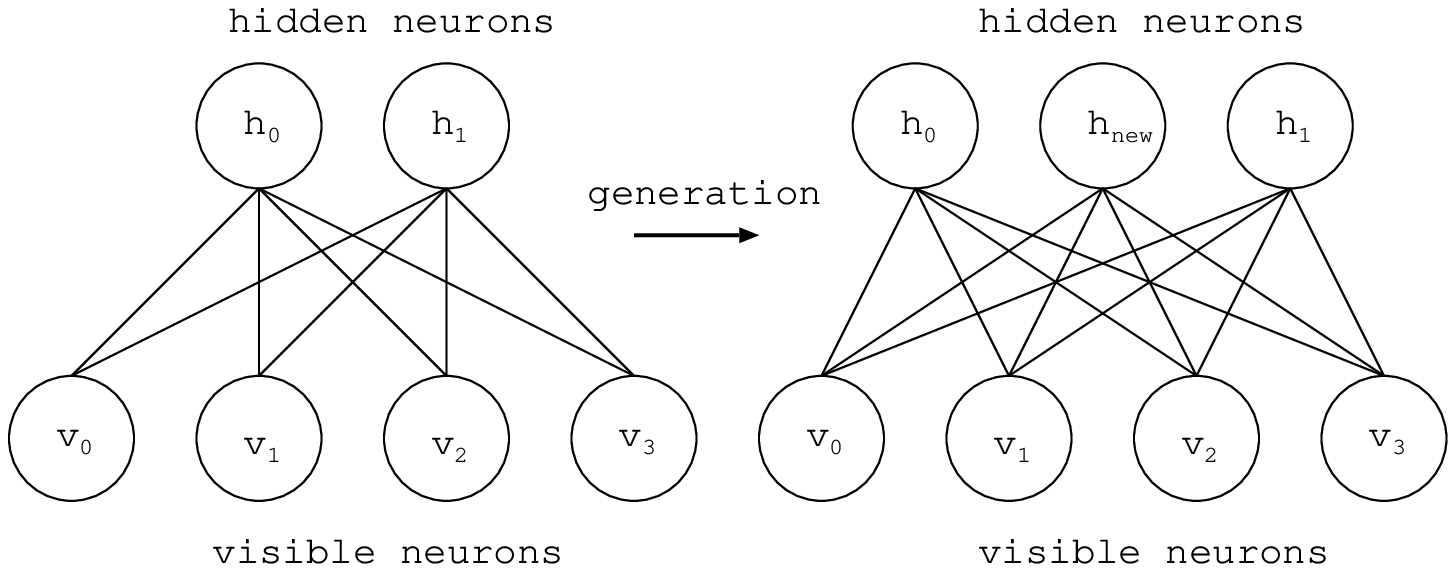}\label{fig:neuron_generation}}
\subfigure[Neuron Annihilation]{\includegraphics[scale=0.5]{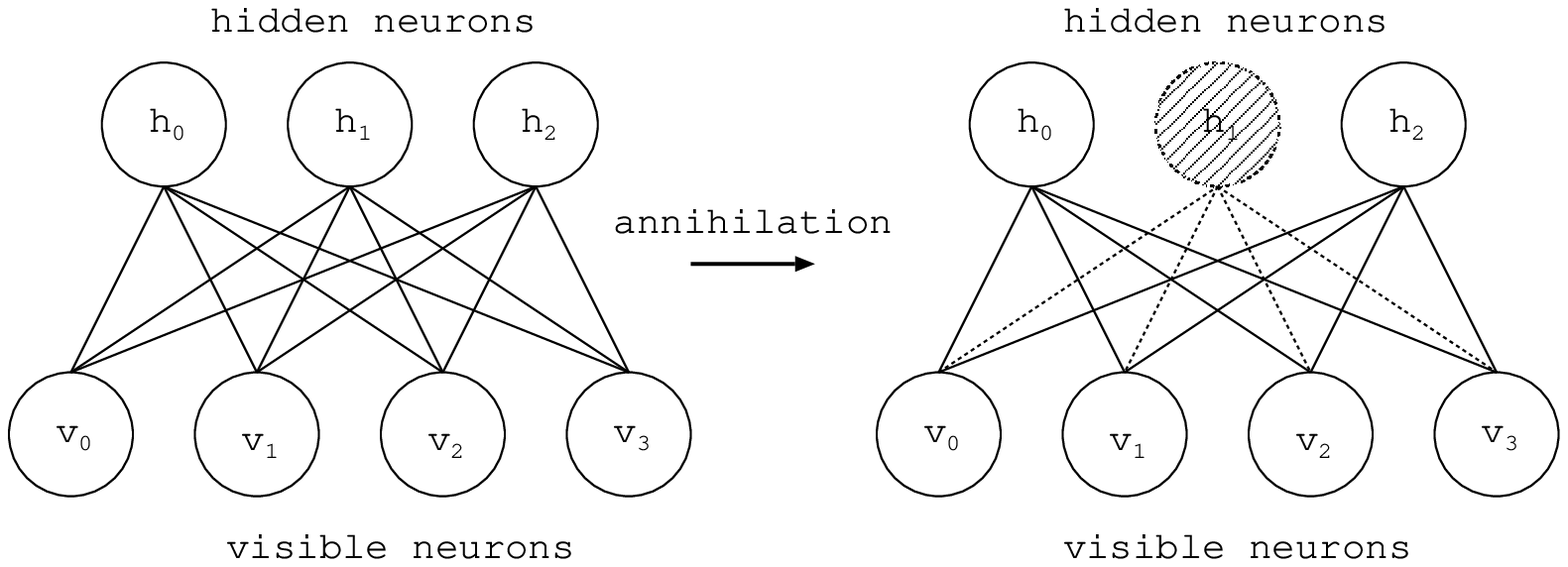}\label{fig:neuron_annihilation}}
%\vspace{-3mm}
\caption{Adaptive RBM}
\label{fig:adaptive_rbm}
\end{center}
\end{figure}

We have already proposed the adaptive structural learning method of RBM (Adaptive RBM) by self-organizing the network structure for given input data \cite{Kamada16_SMC, Kamada16_ICONIP, Kamada16_TENCON}. The method improved the problem that the traditional RBM cannot change its network structure of hidden neurons during the training. The neuron generation / annihilation algorithm, as shown in Fig.\ref{fig:adaptive_rbm}, can determine the suitable number of hidden neurons by monitoring the variance of the parameters during the training. 

The key algorithm monitors the network training situation with the fluctuation of weight vector called the Walking Distance (WD) \cite{Ichimura04}. If a neural network does not have enough neurons to be satisfied to infer, then WD will tend to fluctuate greatly even after a certain period of the training process, because some hidden neurons may not represent an ambiguous pattern due to the lack of the number of hidden neurons. In such a case, we can solve the problem by dividing a neuron which tries to represent the ambiguous patterns into two neurons by inheriting the attributes of the parent hidden neuron. Then we use the condition of neuron generation with inner product of the variance of monitoring two kinds of parameters $\bvec{c}$ and $\bvec{W}$ except $\bvec{b}$ as follows. The reason for the exclusion of $\bvec{b}$ is that the parameter was observed the oscillation according to the input patterns, because the input signals will include some noise data.

\begin{equation}
\label{eq:neuron_generation}
%\vspace{-3mm}
WD_{c_j} \cdot WD_{W_{j}} > \theta_{G},
\end{equation}
\vspace{-3mm}
\begin{eqnarray}
\label{eq:WD_c}
WD_{c_j} &=&  \gamma_{c} WD_{c_j} \nonumber\\
&&+ (1 - \gamma_{c}) (|c_j[\tau] - c_j[\tau-1]|),\\
\label{eq:WD_W}
WD_{W_j} &=&  \gamma_{W} WD_{W_j} \nonumber\\
&&+ (1 - \gamma_{W}) Met(\bvec{W}_j[\tau], \bvec{W}_j[\tau-1]),
\end{eqnarray}
where $WD_{c_{j}} (\geq 0)$ and $WD_{W_{j}}(\geq 0)$ are the variances for a hidden neuron $j$ during training. $WD_{c_j}$, which are calculated by the idea in \cite{Ichimura04}. $\theta_{G} ( > 0)$ is a threshold, which is a certain small value. The neuron generation process has been often occurred if $\theta_G$ is a smaller value. A new hidden neuron $h^{new}_{j} $ with parameters $c^{new}_{j}$ and $W^{new}_{ij}$ will be generated by Eq.(\ref{eq:new_neuron}), if the neuron generation condition is satisfied during the training. Then it will be inserted into the neighborhood of the monitored neuron as shown in Fig.~\ref{fig:neuron_generation}. The initial structure of RBM should be set arbitrary neurons according to the input data set before training.
\begin{equation}
\label{eq:new_neuron}
%\vspace{-3mm}
c^{new}_{j} = c_{j} + N(\mu, \sigma^{2}), \ W^{new}_{ij} = W_{ij} + N(\mu, \sigma^{2}),
\end{equation}
where $N(\mu, \sigma^{2})$ is a random value generated from a normal distribution with the average $\mu$ and the standard deviation $\sigma$. The aim of the addition of the random value to the generated neuron is to prevent to make a distinct of original neuron and generated neuron from learning the same feature of the monitored neuron. In other words, the generated neuron becomes to learn a slightly different feature by adding the noise. To fluctuate the network with a small noise, we set $\mu = 0$ and $\sigma = 0.1$ in this paper.

The annihilation condition for the activation is observed by Eq.(\ref{eq:neuron_annihilation}). The corresponding neuron will be annihilated as shown in Fig.~\ref{fig:neuron_annihilation} if Eq.(\ref{eq:neuron_annihilation}) is satisfied during the training.
\begin{equation}
\label{eq:neuron_annihilation}
%\vspace{-3mm}
\frac{1}{N}\sum_{n=1}^{N} p(h_j = 1 | \bvec{v}_{n}) < \theta_{A},
\end{equation}
where $p(h_j = 1 | \bvec{v}^{(n)})$ means a conditional probability of $h_j \in \{ 0, 1 \}$ for given input data $\bvec{v}^{(n)}$. $\theta_{A}$ is a threshold in $[0, 1]$. The neuron annihilation process has been often occurred if $\theta_A$ is a higher value. 

\subsection{DBN's Layer Generation}
\label{subsec:adaptive_dbn}
We proposed the adaptive structural learning method of DBN (Adaptive DBN) that can determine an optimal number of hidden layers \cite{Kamada16_TENCON}. A RBM employs the adaptive structural learning method by the neuron generation algorithm as described in section \ref{subsec:adaptive_rbm}. In general, data representation of DBN performs the specified features from an abstract concept to concrete representation at each layer in the direction to output layer. That is, a lower layer has the power of non figurative representation, and a higher layer constructs the object to figure out an image of input patterns. Adaptive DBN can automatically adjust an optimal network structure by self-organization.

In the learning process of Adaptive DBN, we observe the $WD$ by left side of Eq.(\ref{eq:neuron_generation}) and the energy function at each RBM layer. If the overall $WD$ and the energy function are still large values, then a new RBM is required to express the suitable network structure for the given input data. That is, the large $WD$ and the large energy function mean that the RBM has lacked data representation capability to figure out an image of input patterns for the teaching signals in the overall network. Therefore, the conditions of layer generation with the total $WD$ and the energy function are defined as Eq.(\ref{eq:layer_generation1}) and Eq.(\ref{eq:layer_generation2}).
\begin{equation}
\sum_{l=1}^{k} (\alpha_{WD} \cdot WD^{l}) > \theta_{L1},
\label{eq:layer_generation1}
\end{equation}
\begin{equation}
\sum_{l=1}^{k} (\alpha_{E} \cdot E^{l}) > \theta_{L2},
\label{eq:layer_generation2}
\end{equation}
where $WD^{l}$ is calculated by left side of Eq.(\ref{eq:neuron_generation}) in the $l$-th RBM. $E^{l}$ is the energy function by Eq.(\ref{eq:energy}) in the $l$-th RBM. $k$ means the current layer. $\alpha_{WD}$ and $\alpha_{E}$ are the constant values for the adjustment of $WD^{l}$ and $E^{l}$, respectively. $\theta_{L1}$ and $\theta_{L2}$ are thresholds, which are small values. Compared with the $\theta_{G}$ in the neuron generation, $\theta_{L1}$ and $\theta_{L2}$ tend to have large values. This means that a new layer will not be likely generated compared with the phenomenon of neuron generation since the layer annihilation is not implemented to the adaptive structural learning method of DBN. If Eq.(\ref{eq:layer_generation1}) and Eq.(\ref{eq:layer_generation2}) are satisfied simultaneously in the learning phase at the layer $k$, a new hidden layer $k+1$ will be generated after the learning process at the layer $k$ is finished.

\section{Knowledge Extraction from Adaptive DBN}
\label{sec:extract_knowledge}
A neural network has been a black box in the sense that while it can approximate any functions, training its structure cannot give any insights on the structure of the approximated functions. Deep learning methods remain such a problem, although various types of them can reach higher classification capability.

For simple machine learning models, for example, SVM, we can see inside of the model, because the model is a weighted linear combination of the feature. However, the deep learning models are some non-linear combination of many neurons, each of which has some parameters including weights. If we have an effective knowledge extraction method, the explicit knowledge, such as IF-Then rule, can be expressed the data representation to infer from the trained deep neural networks with tremendously higher classification capability.

We investigated the signal patterns from the lower layer (close to input) to the upper layer (close to output) for given input data. The signal pattern consisted of input pattern and the calculated pattern in the trained DBN. The relation between input and output was explored by C4.5.

We proposed the Fine Tuning method for the knowledge acquisition from the trained DBN \cite{Kamada16_IWCIA, Kamada17_IJCIS} on the basis of the above mentioned idea. The method is utilizing that the hidden neurons of RBM are represented as binary patterns \{0, 1\}. All possible combinations of binary pattern \{0, 1\} are given to the trained DBN and the signal flow of the network from the lower layer to the upper layer is analyzed the activated path of network. The fine tuning method can repair the activated path of network partially for improving the classification accuracy. The basic idea of the method finds the activated path for erroneous test cases, and the forcibly correct it. With this method, unknown data not seen in the training can be corrected (patched). {\bf Algorithm \ref{alg:finetuning-org}} shows the procedure of the fine tuning method. For $l$-th layer, weights which output wrong patterns are modified by {\bf Algorithm \ref{alg:finetuning-l}}.

In this paper, the explicit IF-THEN rules for the trained DBN were extracted by C4.5. C4.5 is a well known and popular method to generate a decision tree. Of course, there are many decision tree algorithms such as ID3 or C5.0. Among these algorithms, generated rule set of C4.5 is simple one with high classification performance \cite{Badr14}. Therefore, we decided to use C4.5 method to make decision tree for knowledge discover from trained neural network. The extracted rules represent relation between input and output patterns of the Adaptive DBN. {\bf Algorithm \ref{alg:extract_rule}} is the procedure of the rule extraction by C4.5. Fig.~\ref{fig:c45_data_example} is an example of input data of C4.5. As shown in {\bf Algorithm \ref{alg:extract_rule}}, ``x\_p\_i'' is the i-th element in the p-th input pattern. ``y\_p'' is the p-th output pattern.

\begin{algorithm}[tb]
\caption{Fine Tuning Method}
\label{alg:finetuning-org}                        
\begin{algorithmic}[1]
\STATE The $L(1 \leq l \leq L)$ layers network structure by training Adaptive DBN was constructed.
\STATE Fine tuning method described in {\bf Algorithm \ref{alg:finetuning-l}} at the $l$-th layer was implemented from the first layer to the $L$-th layer.
\end{algorithmic}
\end{algorithm}

\begin{algorithm}[tb]
\caption{Procedure of modifying weight at the $l$-th layer}
\label{alg:finetuning-l}                        
\begin{algorithmic}[1]
 \STATE Let $X(\bvec{x}_{1}, \cdots, \bvec{x}_p ,\cdots, \bvec{x}_{N})$ be input patterns, where $N$ is the number of patterns. $\bvec{x}_{p}$ is a $I$-dimensional vector. Let $Y(y_{1}, \cdots, y_p ,\cdots, y_{N})$ be teacher signals. The trained DBN has $L(1 \leq l \leq L)$ layers. Let $l$ is the target layer for fine tuning.

\STATE Give $X$ to the trained network for feed forward calculation. In the calculation, save information which neuron was activated from lower to upper layer. Let $X^{T}$ and $X^{F}$ be correct and wrong input patterns for $Y$, respectively.

\STATE In the $l$-th layer, if a neuron $j$ is satisfied with Eq.(\ref{eq:condition_A}), modify weights connected to the neuron $j$ to $w^{correct}$. Eq.(\ref{eq:condition_A}) is the ratio that only $X^{T}$ is fired at the neuron $j$. $w^{correct}$ is a constant value (e.g. $w^{correct} = 1$).
\begin{equation}
\label{eq:condition_A}
 \frac{|Act^{T}_{j}|}{|X^{T}| + |X^{F}|} \geq \theta^{T},
\end{equation}
where, $\theta^{T} (0 \leq \theta^{T} \leq 100)$ is the threshold value. 

\STATE In the $l$-th layer, if a neuron $j$ is satisfied with Eq.(\ref{eq:condition_B}), modify weights connected to the neuron $j$ to $w^{wrong}$. Eq.(\ref{eq:condition_B}) is the ratio that only $X^{F}$ is fired at the neuron $j$. $w^{wrong}$ is a constant value (e.g. $w^{wrong} = 0$).
\begin{equation}
\label{eq:condition_B}
\frac{|Act^{F}_{j}|}{|X^{T}| + |X^{F}|} \geq \theta^{F},
\end{equation}
where, $\theta^{F} (0 \leq \theta^{F} \leq 100)$ is the threshold value.

\end{algorithmic}
\end{algorithm}

\begin{algorithm}[tbp]
\caption{Knowledge Extraction by C4.5}
\label{alg:extract_rule}                        
\begin{algorithmic}[1]
\STATE Let $X(\bvec{x}_{1}, \cdots, \bvec{x}_p ,\cdots, \bvec{x}_{N})$ be input patterns, where $N$ is the number of patterns. $\bvec{x}_{p}$ is a $I$-dimensional vector. The trained DBN has $L(1 \leq l \leq L)$ layers. Let $Net()$ be the function which calculates output patterns $Y(y_{1}, \cdots, y_p, \cdots, y_{N})$ for given $X$. $y_{p}$ has a class value for classification.
\STATE Give $X$ to $Net()$ and calculate $Y$.
\STATE Extract rules by C4.5, where input is $X$, teacher signal is $Y$ (Fig.~\ref{fig:c45_data_example} is an example of input-output patterns for C4.5.).
\end{algorithmic}
\end{algorithm}

\begin{figure}[tbp]
\centering
\begin{small}
\begin{lstlisting}[numbers=none]
x_1_1, x_1_2, x_1_3, ..., x_1_I, y_1
x_2_1, x_2_2, x_2_3, ..., x_2_I, y_2
x_3_1, x_3_2, x_3_3, ..., x_3_I, y_3

...

x_p_1, x_p_2, x_p_3, ..., x_p_I, y_p

...

x_N_1, x_N_2, x_N_3, ..., x_N_I, y_N
\end{lstlisting}
\end{small}
\caption{An example of input and output patterns for C4.5}
\label{fig:c45_data_example}
\end{figure}

\begin{table*}[tbp]
\caption{Blood test items}
\label{tab:health_check_items}
\vspace{-3mm}
\centering
\begin{tabular}{c}
% 1
\begin{minipage}[t]{0.5\hsize}
\scalebox{0.8}[0.8]{
\begin{tabular}[t]{l|l|l|l}
\hline \hline
Category&Name&Data type&Range\\ \hline \hline
Basic test&Patient ID&Integer&\\ \cline{2-4} 
&Age&Integer&(10 - 134)\\ \cline{2-4} 
&Sex&Code&[Male, Female]\\ \cline{2-4} 
&Date&Integer&\\ \cline{2-4} 
&Height&Float&(117 - 196.7)\\ \cline{2-4} 
&Weight&Float&(27.6 - 175)\\ \cline{2-4} 
&BMI&Float&(11.9 - 57.3)\\ \cline{2-4} 
&Abdomen&Float&(53 - 157)\\ \cline{2-4} 
&Eye sight(right)&Float&(0 - 9.915)\\ \cline{2-4} 
&Eye sight(left)&Float&(0 - 9.915)\\ \cline{2-4} 
&Hearing(right, 1000)&Code&[Normal, Abnormal]\\ \cline{2-4} 
&Hearing(right, 4000)&Code&[Normal, Abnormal]\\ \cline{2-4} 
&Hearing(left, 1000)&Code&[Normal, Abnormal]\\ \cline{2-4} 
&Hearing(left, 4000)&Code&[Normal, Abnormal]\\ \hline
Blood pressure&Blood pressure(Max)&Integer&(70 - 257)\\ \cline{2-4} 
&Blood pressure(Min)&Integer&(26 - 148)\\ \hline
Urine&Protein&Code&[(−),(±),(1+),(2+),(3+) ]\\ \cline{2-4} 
&Occult blood&Code&[(−),(±),(1+),(2+),(3+) ]\\ \cline{2-4} 
&Urobilinogen&Code&[(−),(±),(1+),(2+),(3+) ]\\ \hline
Blood analysis&WBC&Integer&(1200 - 26000)\\ \cline{2-4} 
&RBC&Integer&(234 - 672)\\ \cline{2-4} 
&Hb&Float&(5.5 - 22.3)\\ \cline{2-4} 
&Ht&Float&(20.6 - 65.2)\\ \cline{2-4} 
&PLT&Float&(2.7 - 112.6)\\ \hline
Lipid&LDL&Integer&(4 - 357)\\ \cline{2-4} 
&HDL&Integer&(17 - 205)\\ \cline{2-4} 
&TG&Integer&(17 - 2628)\\ \cline{2-4} 
&Sugar urine&Code&[(−),(±),(1+),(2+),(3+) ]\\\hline
Diabetes&Blood sugar&Integer&(41 - 441)\\ \cline{2-4} 
&HbA1c&Float&(4.7 - 12.7)\\ \cline{2-4} 
&Uric acid&Integer&[1, 3, 4, 6, 7]\\ 
\hline \hline
\end{tabular}
} 
\end{minipage}
% 2
\begin{minipage}[t]{0.5\hsize}
\scalebox{0.8}[0.8]{
\begin{tabular}[t]{l|l|l|l}
\hline \hline
Category&Name&Data type&Range\\ \hline \hline
Liver function&GOT&Integer&(5 - 1134)\\ \cline{2-4} 
&GPT&Integer&(4 - 1909)\\ \cline{2-4} 
&Gamma GTP&Integer&(4 - 2329)\\ \cline{2-4} 
&ALP&Integer&(39 - 1758)\\ \cline{2-4} 
&LDH&Integer&(77 - 620)\\ \cline{2-4} 
&ChE&Integer&(103 - 621)\\ \cline{2-4} 
&ZTT&Float&(1 - 43.8)\\ \cline{2-4} 
&Total Bilirubin&Float&(0.1 - 4.8)\\ \cline{2-4} 
&TP&Float&(5.6 - 9.3)\\ \cline{2-4} 
&Alb&Float&(3.3 - 5.4)\\ \cline{2-4} 
&A/G&Float&(0.6 - 2.8)\\ \cline{2-4} 
&TC&Integer&(94 - 418)\\ \hline
Uric acid&Creatinine&Float&(0.28 - 13.68)\\ \hline
Kidney function&BUN&Integer&(5 - 59)\\ \cline{2-4} 
&eGFR&Float&(3.9 - 224.7)\\ \cline{2-4} 
&CRP&Code&[(−),(±),(1+),(2+),(3+)]\\ \hline
Infection&Hbs antigen&Code&[(−),(＋) ]\\ \cline{2-4} 
&Hbs antibody&Code&[(−),(＋) ]\\ \cline{2-4} 
&Hbc antibody&Code&[(−),(＋)]\\ \cline{2-4} 
&Hcv antibody&Code&[(−),(±),(1+),(2+)]\\ \cline{2-4} 
&Pepsinogen&Code&[(−),(＋)]\\ \hline
Additional test&Pylori&Code&[(−),(＋)]\\ \cline{2-4} 
&Amylase&Float&(27 - 1335)\\ \cline{2-4} 
&ASO&Float&(10 - 393)\\ \cline{2-4} 
&CEA&Float&(0.2 - 10)\\ \cline{2-4} 
&CA15-3&Float&(4.600-19.400)\\ \cline{2-4} 
&TTT&Float&(0.3 - 10.6)\\ \cline{2-4} 
&Fecal occult blood&Code&[(−),(＋) ]\\ \hline
Other&Health questionnaire&Code&[1, 2]\\
\hline \hline
\end{tabular}
} 
\end{minipage}
\end{tabular}
%\vspace{-5mm}
\end{table*}

\section{Experiment}
\label{sec:exe}
\subsection{Prediction System for Health care Condition}
\label{subsec:exe_cancer_prediction}
In this section, the effectiveness of our proposed Adaptive DBN was verified on the medical examination data in order to develop the prediction system of cancer. For CIFAR-10 \& CIFAR-100 \cite{CIFAR10}, the classification capability was  97.4\% and 81.2\% for test cases, respectively. The accuracy ratio is higher than the traditional DBN\cite{Kamada16_TENCON,Kamada17_IWCIA}. We challenge to apply our proposed method to medical database as the complex big data.

The medical data is a data set for health check which was provided from `Hiroshima Environment and Health Association \cite{Kanhokyo}'. The 100,000 records were collected from April 2012 to March 2015. Each record consists of 59 items including blood tests as shown in Table~\ref{tab:health_check_items} and four kinds of medical images as shown in Table~\ref{tab:medicalimage_type}. In this paper, two kinds of medical images, `Lung' cancer and `Stomach' cancer were used for the cancer prediction. Each medical image is judged normal or abnormal for the class label by medical expert. The 100,000 examination data were randomly divided into 80,000 training cases and 20,000 test cases in the proportion of four to one.

The following parameters were used for the training of our Adaptive DBN. The training algorithm was Stochastic Gradient Descent (SGD) method, the batch size was 100, the learning rate was 0.005, the initial number of hidden neurons was 400, $\theta_G = 0.001, \theta_A = 0.100$, $\theta_{L1} = 0.05$, and $\theta_{L2} = 0.05$. The two sets, the blood tests and the Lung images, and the blood tests and the Stomach image were trained by the Adaptive DBN. To compare the performance of the proposed Adaptive DBN with the other methods, `Linear SVM', `Traditional RBM', and `Traditional DBN' were used in the experiment. The SVM was used to grasp the statistical overview of the medical database instead of CNNs. Although CNN is a popular deep learning method, it was not used in the experiment because of the problem for an optimal network design (e.g. the setting of layers, filters, and so on.).

\begin{table}[tbp]
\caption{Medical image}
\vspace{-5mm}
\label{tab:medicalimage_type}
\begin{center}
\scalebox{0.8}[0.8]{
\begin{tabular}{l|l|l|l}
\hline \hline
Type & Body part & Method & Cancer test\\ \hline \hline
X-ray for lung & Lung & X-ray & \\ \hline
CT for lung & Lung & CT & $\bigcirc$ \\ \hline
X-ray for stomach & Stomach & X-ray & $\bigcirc$\\ \hline
Mammography & Breast & Mammography (X-ray)& $\bigcirc$ \\ \hline
\hline 
\end{tabular}
} 
\end{center}
%\vspace{-5mm}
\end{table}

\begin{figure}[tbp]
\centering
\subfigure[Lung cancer]{
  \includegraphics[scale=0.6]{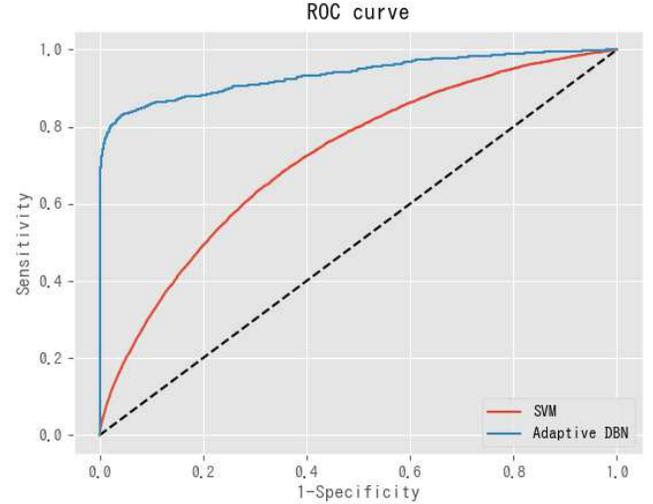}
  \label{fig:roc_chest}
}
\subfigure[Stomach cancer]{
  \includegraphics[scale=0.6]{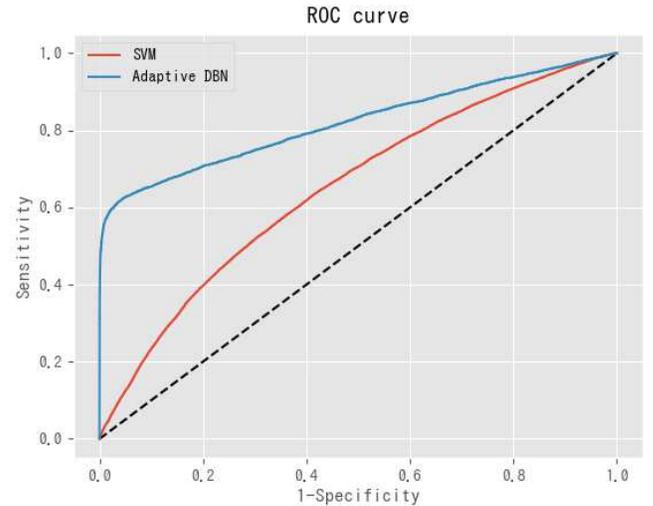}
  \label{fig:roc_stomach}
}
%\vspace{-3mm}
\caption{ROC Curve}
\label{fig:roc}
%\vspace{-3mm}
\end{figure}

Table~\ref{tab:accuracy_medical_data} shows the average value and the standard deviation of the classification accuracy on 10-trials for the test data in the medical data. The classification accuracy of the Adaptive DBN was 95.5\% for Lung cancer, 94.3\% for Stomach cancer, which were higher values than not only the traditional RBM and DBN but also SVM. By using the fine tuning method of {\bf Algorithm \ref{alg:finetuning-org}} and {\bf Algorithm \ref{alg:finetuning-l}}, the classification accuracy of the Adaptive DBN was improved to 98.1\% for Lung cancer, 98.0\% for Stomach cancer. Fig.~\ref{fig:roc} shows the ROC (Receiver Operating Characteristic) curve for two kinds of cancers. The ROC curve plots sensitivity (true positive rate) on vertical axis against 1-specificity (false positive rate) on horizontal axis. The Adaptive DBN shows good performance since the area under the curve of the Adaptive DBN was larger than the SVM. %By using the result of the Adaptive DBN, the interface system which can show the probability of cancers for a given input was developed as shown in Fig.~\ref{fig:predictionsystem_android}. The system can give attention to the doctor when the patient seems to be a tendency of pathological abnormal. Even when there is no expert medical practitioner such as in the island, this system can point out the requirement of more precise examination and of follow-up observation.

\begin{table}[tb]
\caption{Classification Accuracy for the medical data}
%\vspace{-3mm}
\label{tab:accuracy_medical_data}
\centering
%\scalebox{0.8}[0.8]{
\begin{tabular}{l|r|r|r|r}
\hline \hline
&\multicolumn{4}{|c}{Accuracy}\\ \cline{2-5}
&\multicolumn{2}{|c}{Lung} & \multicolumn{2}{|c}{Stomach} \\ \cline{2-5}
\multicolumn{1}{c|}{Model} &\multicolumn{1}{c|}{Ave.} & \multicolumn{1}{c|}{Std.} & \multicolumn{1}{c|}{Ave.} &\multicolumn{1}{c}{Std.}\\ \hline
\hline
Traditional RBM &0.833 & 0.011 &0.833 & 0.011 \\ \hline
Adaptive RBM  & {\bf 0.854} & 0.011  & {\bf 0.854} & 0.011 \\ \hline
Traditional DBN & 0.911 & 0.010 & 0.911 & 0.010 \\ \hline
Adaptive DBN  & {\bf 0.955} & 0.010 & {\bf 0.943} & 0.010 \\ \hline
\hline
SVM  &  0.696 & 0.005 &  0.696 & 0.006 \\ \hline
\hline
\end{tabular}
%} 
\end{table}

\begin{figure}[tbp]
\centering
\subfigure[Lung cancer]{
  \includegraphics[scale=0.4]{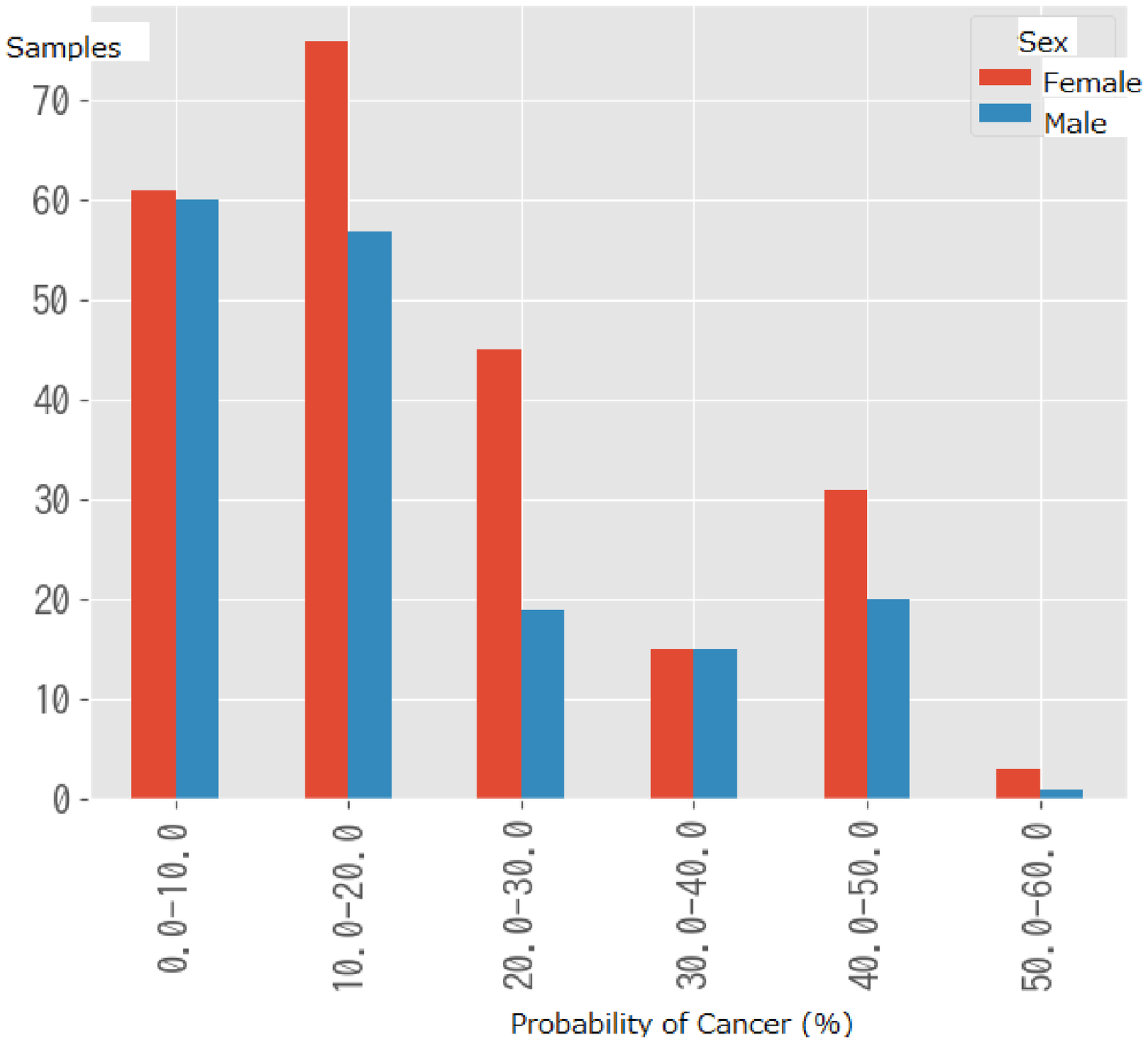}
  \label{fig:chest_x_y}
}
\subfigure[Stomach cancer]{
  \includegraphics[scale=0.4]{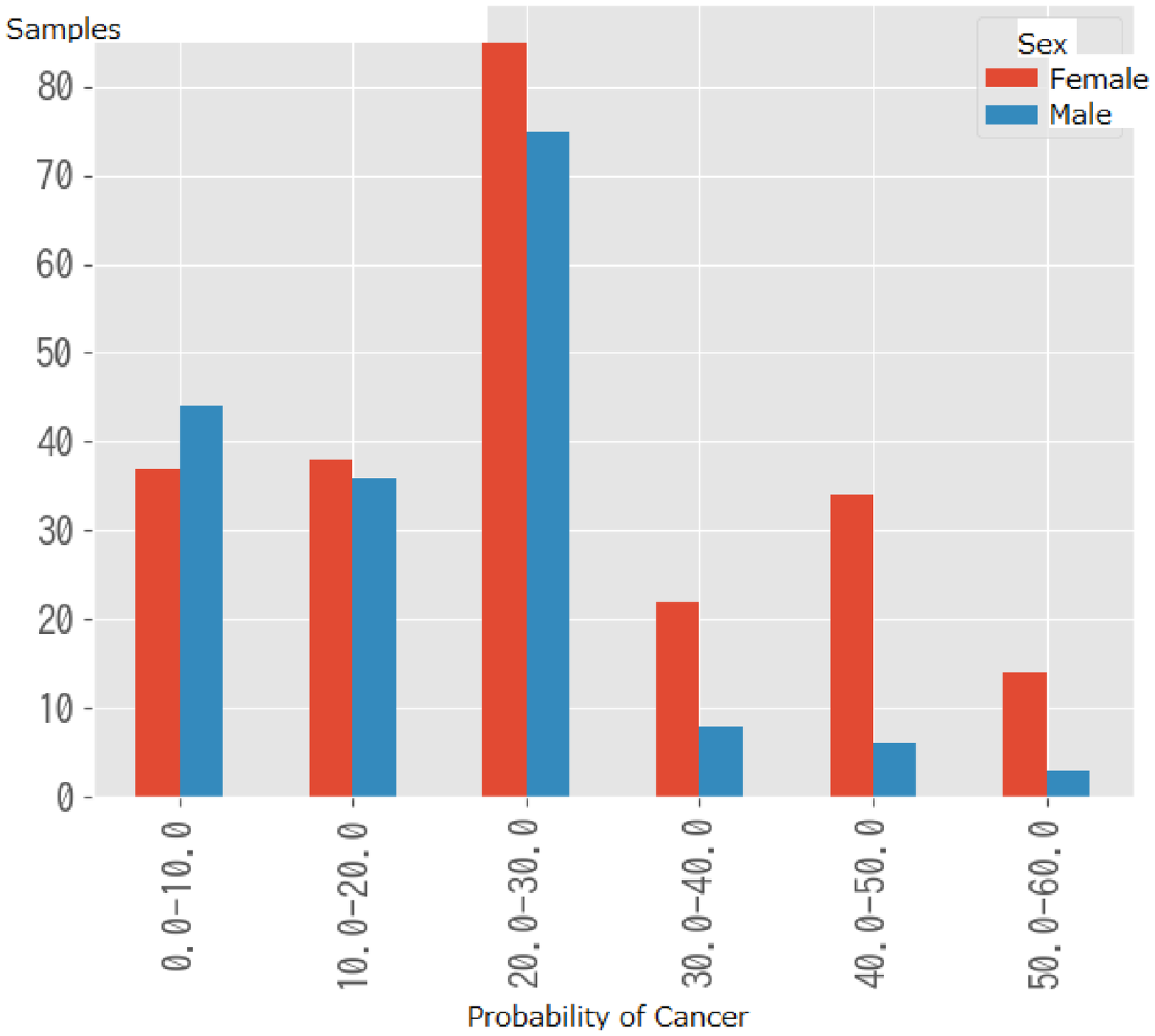}
  \label{fig:stomach_x_y}
}
\vspace{-3mm}
\caption{Histogram of predicted possibility of cancer}
\label{fig:cancer_output}
%\vspace{-3mm}
\end{figure}

\subsection{Knowledge Extraction from the trained Adaptive DBN}
\label{subsec:exe_rule}
The Adaptive DBN can output a possibility of two kinds of cancers from the blood tests and Lung and Stomach images. In this section, the explicit knowledge represented as IF-THEN rules was extracted from the trained Adaptive DBN by C4.5. About 400 cases of the blood tests without medical image were used to evaluate the extraction method, because the 400 subjects did not receive CT or MRI examination. Each record of the blood tests has same items as Table~\ref{tab:health_check_items}. Because the data has only blood tests, the pair of the blood tests and averaged value of medical image (stereotyped data not including disease) was given to the trained Adaptive DBN, and the probability for each cancer was calculated. Fig.\ref{fig:cancer_output} shows the histogram of the predicted possibility for two kinds of cancers. For example, there was no case that the possibility of cancer is more than 60\% for two kinds of cancer. Based on {\bf Algorithm \ref{alg:extract_rule}}, the pair of input (blood tests) and output for cancer with more than 20\% probability, was analyzed by C4.5. The probability ratio is not high, but it is remarkable value to infer the suspected cancer cases.

%\begin{figure}[tbp]
%\centering
%\includegraphics[scale=0.7]{fig/prediction_system_android.eps}
%\vspace{-3mm}
%\caption{Prediction system for medical examination}
%\label{fig:predictionsystem_android}
%\end{figure}

\begin{figure}[tb]
\centering
\begin{small}
\begin{lstlisting}[numbers=none]
gamma_gtp <= 78 :
|   egfr <= 92 :
|   |   age <= 48 :
|   |   |   got <= 17 : 5 (9.0/1.0)
|   |   |   got > 17 : 3 (25.0/1.0)
|   |   age > 48 :
|   |   |   plt > 17 : 3 (250.0)
|   |   |   plt <= 17 :
|   |   |   |   tg <= 162 : 3 (57.0/1.0)
|   |   |   |   tg > 162 : 5 (3.0)
|   egfr > 92 :
|   |   plt <= 18 : 5 (4.0)
|   |   plt > 18 :
|   |   |   tp > 7 : 5 (3.0/1.0)
|   |   |   tp <= 7 :
|   |   |   |   urea_nitrogen > 10 : 3 (18.0/1.0)
|   |   |   |   urea_nitrogen <= 10 :
|   |   |   |   |   uric_acid <= 4 : 3 (4.0/1.0)
|   |   |   |   |   uric_acid > 4 : 6 (2.0)
gamma_gtp > 78 :
|   age <= 50 : 6 (10.0)
|   age > 50 :
|   |   plt <= 18 : 5 (3.0)
|   |   plt > 18 : 3 (13.0)
\end{lstlisting}
\end{small}
\caption{An example of decision tree tree in C4.5}
\label{fig:c45_tree_example}
\end{figure}
 
Fig.~\ref{fig:c45_tree_example} shows a part of the extracted decision tree of C4.5. Table~\ref{tab:c45_rule} shows the extracted 18 IF-THEN rules. The inference by using the extracted rules becomes 98.2\% accuracy for the test data and the ratio seems to be high ratio, although the ratio was not better than the results of DBN. 

From the Table~\ref{tab:c45_rule}, there was totally a tendency for abnormality to be observed in `WBC (white blood cell)' in case that the possibility of cancer was high. For the lung cancer, abnormalities related to Albumin and Total protein were found. For the stomach cancer, abnormalities related to GPT, GOT and $\gamma$GTP were found. Although these items are not directly related with the cancers, such cases are actually reported on \cite{alb, a6}. For example, it is said that the ratio of the protein in blood is decreased as a cancer progresses. The evaluation by using another data should be required, but the extracted rules will be medically considered knowledge.

\begin{table*}[tbp]
\caption{Extracted Knowledge from the trained Adaptive DBN by using C4.5}
%\vspace{-5mm}
\label{tab:c45_rule}
\begin{center}
%\scalebox{0.9}[0.9]{
\begin{tabular}{c|c|p{7cm}|l}
\hline \hline
Rule ID & Cancer & Condition & Cancer Feature\\ \hline\hline
A-1  & Lung & Age is less than 50, AND WBC is within normal range. & Possibility of cancer is lower than 20\%. \\ \cline{1-3}
A-2  & Lung & Age is more than 50, AND WBC, BMI, Hb, ALP, and eGFR are with in normal range. &\\ \hline
A-3  & Lung & WBC is out of normal range. & Possibility of cancer is higher than 20\% and lower than 35\%. \\ \hline
A-4  & Lung & WBC, Albumin, and Total protein are out of normal range. & Possibility of cancer is higher than 35\% and lower than 50\%. \\ \hline
A-5  & Lung & WBC and RBC are out of normal range. & Possibility of cancer is higher than 50\%.  \\ \cline{1-3}
A-6  & Lung & Age is higher than 80 AND WBC, BMI, ALP, Albumin, Hb, and eGFR are out of normal range. &  \\ \hline
\hline
B-1  & Stomach & Age is lower than 50 AND WBC is with in normal range. &  Possibility of cancer is lower than 20\%. \\ \cline{1-3}
B-2  & Stomach & Age is higher than 50, AND BMI, Albumin, ALP, eGFR are with in normal range. & \\ \hline
B-3  & Stomach & WBC is out of normal range. & Possibility of cancer is higher than 20\% and lower than 35\%. \\ \cline{1-3}
B-4  & Stomach & WBC, GPT, and GOT are out of normal range. &   \\ \cline{1-3}
B-5  & Stomach & WBC and $\gamma$GTP are out of normal range. &  \\ \hline
B-6  & Stomach & WBC and ASO are out of normal range. & Possibility of cancer is higher than 35\% and lower than 50\%. \\ \cline{1-3}
B-7  & Stomach & WBC, GPT, GOT, $\gamma$GTP, and Creatinine are out of normal range. &  \\ \hline
B-8  & Stomach & Age is more than 80, AND WBC, BMI, HbA1c albumin, and eGFR are out of normal range. & Possibility of cancer is higher than 50\%. \\ \cline{1-3}
B-9  & Stomach & WBC and Pylori are out of normal range. &     \\ \cline{1-3}
B-10 & Stomach & WBC and CA19\_9 are out of normal range. &    \\ \cline{1-3}
B-11 & Stomach & WBC and Amylase are out of normal range. &  \\ \cline{1-3}
B-12 & Stomach & WBC and Elastase are out of normal range. &  \\ \hline
\hline 
\end{tabular}
%} 
\end{center}

\end{table*}

\section{Conclusion}
\label{sec:conclusion}
The adaptive structural learning method of DBN which shows higher classification accuracy than traditional method, has been the expected method for Deep Learning. The method was applied to medical examination data and the medical system to give a doctor information related to the abnormal or normal of cancer was developed. In this paper, the explicit knowledge represented IF-THEN rules was extracted from the training result of the Adaptive DBN by using C4.5. As a result, some interesting features to find a cancer at an early stage were acquired. These abnormalities in only blood test items can not be said cancer, but the extracted rules will be medically considered knowledge. To show the effectiveness of our method, verification of the extracted rules with medical experts is required in the future. The trained Adaptive DBN have already showed high classification accuracy for the medical data. We will need to investigate the difference of output space between the Adaptive DBN and C4.5.
%verification with more data is required in the future.

\section*{Acknowledgment}
This work was supported by JSPS KAKENHI Grant Number JP17J11178.

\end{document}